\documentclass[letterpaper, 10 pt, conference]{ieeeconf}  

\IEEEoverridecommandlockouts                              

\usepackage{framed,multirow}

\usepackage{amssymb}
\usepackage{latexsym}
\usepackage{bbm}
\usepackage{times}
\usepackage{epsfig}
\usepackage{graphicx}
\usepackage{amsmath}
\usepackage{subfigure}
\usepackage[english]{babel}
\usepackage{booktabs}
\usepackage{lineno}
\usepackage{url}
\usepackage{xcolor}
\usepackage{rotating}

\title{\LARGE \bf
In pixels we trust: From Pixel Labeling to Object Localization \\ and Scene Categorization}

\author{Carlos Herranz-Perdiguero$^{1}$, Carolina Redondo-Cabrera$^{1}$ and Roberto J. L\'opez-Sastre$^{1}$
 \thanks{*This work is supported by project PREPEATE, with reference TEC2016-80326-R, of the Spanish Ministry of Economy, Industry and Competitiveness. We gratefully acknowledge the support of NVIDIA Corporation with the donation of a GPU used for this research. Cloud computing resources were kindly provided through a Microsoft Azure for Research Award.}
 \thanks{$^{1}$The authors are with GRAM research group, Department of Signal Theory and Communications,
         University of Alcal\'a, 28805, Alcal\'a de Henares, Spain
         {\tt\small \{c.herranz,carolina.redondoc\}@edu.uah.es, robertoj.lopez@.uah.es}}%
 }

\def\eg{\emph{e.g. }}

\def\etal{\emph{et al. }}
\def\ie{\emph{i.e. }}

\begin{document}

\maketitle
\thispagestyle{empty}
\pagestyle{empty}

\begin{abstract}
While there has been significant progress in solving the problems of image pixel labeling, object detection and scene classification, existing approaches normally address them separately. In this paper, we propose to tackle these problems from a bottom-up perspective, where we simply need a semantic segmentation of the scene as input. We employ the DeepLab architecture, based on the ResNet deep network, which leverages multi-scale inputs to later fuse their responses to perform a precise pixel labeling of the scene. This semantic segmentation mask is used to localize the objects and to recognize the scene, following two simple yet effective strategies. We evaluate the benefits of our solutions, performing a thorough experimental evaluation on the NYU Depth V2 dataset. Our approach achieves a performance that beats the leading results by a significant margin, defining the new state of the art in this benchmark for the three tasks comprising the scene understanding: semantic segmentation, object detection and scene categorization.
\end{abstract}

\section{INTRODUCTION}
\label{sec:intro}
Visual semantic scene understanding has become a crucial capability in many applications, such as autonomous driving \cite{Ha2017}, indoor navigation \cite{Ma2017} or robotic manipulation \cite{Wong2017}. In fact, it is the capability that enables robots to truly interact with the environment with a minimal intervention or number of sensors, due to the rich information contained in images.

Its goal consists in obtaining as much semantic knowledge of a given scene as possible. This includes scene categorization (labeling the whole scene), object detection (predicting object locations by bounding boxes), and semantic segmentation (labeling each pixel with a category).

In recent years, there has been large progress in solving all these tasks (\eg \cite{Chen2016,Redmon2017,Xiao2010,Espinace2010}). However, most current state-of-the-art approaches propose to solve these problems separately. In this paper, we demonstrate that it results beneficial to address these problems following a bottom-up approach, which mostly relies on a precise semantic segmentation of the scene.

\begin{figure}[t]
\centering
\includegraphics[width=0.9\linewidth]{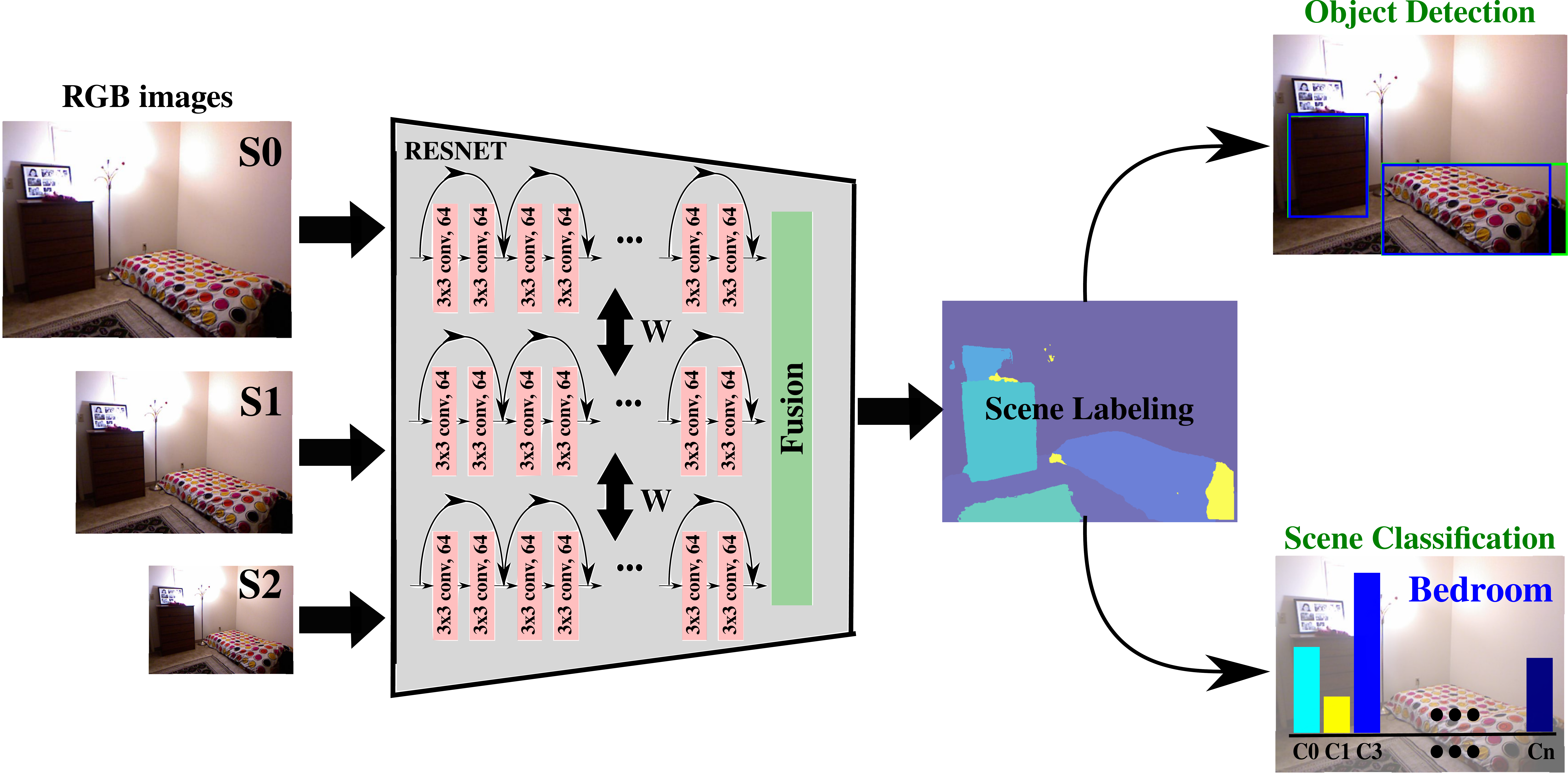}
\caption{We train a multi-scale DeepLab architecture (ResNet based) to perform a precise pixel labeling of the scene. The key innovation is to leverage on this semantic segmentation to perform both the object localization and scene recognition, offering a fast and precise integral solution, which can be embedded in any mobile robotic platform.}
\label{fig:graphical_abstract}
\end{figure}

As it is shown in Figure \ref{fig:graphical_abstract}, our solution starts with a pixel labeling deep network. We base our design on the DeepLab architecture \cite{Chen2016}, using ResNet deep networks \cite{He2016}, enhancing its performance with a multi-scale supervision. We then propose two simple yet precise approaches to perform both the object localization and the scene recognition tasks considering only the scene semantic segmentation as input.

Overall, the main contributions of our work are as follows:
\begin{enumerate}
 \item In principle, the core of the proposed architecture relies on a semantic segmentation network. We follow an intuitive extension of the DeepLab pixel labeling model \cite{Chen2016}, which consists in fusing the response of a pool of multi-scale ResNet based networks.
 \item Starting from our precise semantic segmentation, we demonstrate that it is possible to successfully classify the scenes. Technically, we introduce two classifiers based solutions. First, we design a simple approach that uses SVMs with additive kernels, using as features the histograms of class labels computed from the semantic segmentation. Second, by viewing each semantic segmentation as a one-hot binary vector, indicating the presence/absence of the corresponding classes, we propose to perform the categorization of the scene with a Linear SVM. We show that these two surprisingly simple and fast solutions can surpass prior state-of-the-art results reported by complex deep networks trained to directly classify the scenes.
 \item We finally showcase the potential of our precise semantic segmentations via the task of object localization. We just need to identify in the pixel labeling mask for the whole scene, the connected components that enclose the objects, allowing us to localize them. 
 \item Without bells and whistles, all our models surpass all previous state-of-the-art results on the challenging NYU-Depth V2 (NYUD2) dataset \cite{Silberman2012} for the three tasks comprising scene understanding: scene categorization, object detection and semantic segmentation.
\end{enumerate}

The rest of the paper is organized as follows. Section \ref{relatedwork} reviews the previous works. We technically detail our approaches in Section \ref{approach}. Section \ref{exp} includes the experimental evaluation. We conclude in Section \ref{conclusion}.

\section{RELATED WORK}
\label{relatedwork}

Recently, CNNs have become the state-of-the-art approach in order to solve scene understanding related tasks separately. Increasingly better results have been achieved in semantic segmentation (\eg \cite{Ma2017,Chen2016,Zhao2017}), scene classification (\eg \cite{He2016,Wang2016,Simonyan2015,Song2017}) and object detection (\eg \cite{Redmon2017,Liu2016}).

However, not many works deal with the scene understanding problem \emph{as a whole}. Among those that have done it, we find the earlier works which use handcrafted features, to capture some specific properties considered as representative. Probably, one of the first approaches in trying to get a major scene understanding can be found in \cite{LiJia2009}. They propose a hierarchical generative model that is able to classify the image (scene classification) and localize the objects that are part of it. Moreover, based on the relationship between the type of scene and the objects, the model provides pixel level segmentations (semantic segmentation). In \cite{Yao2012}, a holistic model capable of accomplishing all the three tasks is presented too. They make use of a hierarchical graphical model, a Conditional Random Field, with variables representing aspects such as the presence or absence of a class, the class labels, or the correctness of a certain candidate for object detection. That way, the model can learn jointly about all the different tasks comprising scene understanding. Gupta \etal \cite{Gupta2013,Gupta2015} propose a method to detect contours on depth images for semantic segmentation, and subsequently quantize the segmentation outputs as local features for scene classification. More recently, the use of multi-layered neural networks makes possible to learn features directly from large amounts of data to solve the scene understanding problem. For instance, Gupta \etal \cite{Gupta2014} use the R-CNN \cite{Girshick2014} detector on depth images to localize objects in indoor scenes. Then, the output from their object detectors is used with a superpixel classification framework to obtain a semantic segmentation of the scene.

Our work clearly differs from the references above. First, unlike \cite{Gupta2013, Gupta2014, Gupta2015, Wang2016, Song2017}, our approach does \emph{not} need to use depth images, we solve the whole scene understanding task with just one modality, \ie RGB images only. Second, we base the solution to the problem of scene understanding on an accurate semantic segmentation. Neither contour detectors \cite{Gupta2013,Gupta2015}, nor object detections \cite{Gupta2014} are needed. Instead, we directly obtain the pixel labeling model with a multi-scale deep network. We then solve the tasks of object localization and scene categorization by simply using the semantic segmentation mask as a feature.

\section{SCENE UNDERSTANDING}
\label{approach}

\subsection{Scene Labeling Model}
\label{seg_model}

The use of Fully Convolutional Networks (FCNs), first introduced by Long \etal \cite{Long2015}, has achieved great results by adapting image classification deep networks to the task of semantic segmentation. In a nutshell, FCNs are mainly based on the replacement of the last fully-connected layers with convolutional ones.

Here, we propose the use of such networks, following the state-of-the-art publicly available model for scene labeling known as DeepLab \cite{Chen2016}. This model introduces three mayor contributions: 1) atrous convolutions, as a way of increasing the field of view (FOV) of the convolutional filters; 2) Atrous Spatial Pyramid Pooling (ASPP) technique, which makes use of several parallel atrous convolutions at different rates, to extract features at various scales before merging them; and 3) a post-processing CRF to refine the scene labeling results given by the CNN under the premise that nearby and similar appearance pixels should be given the same label.

Technically, for our scene labeling model, we adopt this DeepLab architecture, incorporating a ResNet-101 based network \cite{He2016} to it. For the indoor scene labeling problem we want to address with the NYUD2 dataset, we follow a joint multi-scale learning procedure. For that purpose, we process the input image with scaling factors \{0.5, 0.75, 1\} of the original resolution. Thereby, we actually have three complete ResNet-based networks working in parallel, as branches of the whole architecture (see Figure \ref{fig:graphical_abstract}). At the end, we fuse the results at each pixel position by taking the maximum response given by the network for each scale. This procedure can be seen as a way of increasing and improving the contextual information that the network obtains from the images. The effect when processing smaller scales is similar to increasing the FOV of the filters since each filter \emph{covers} a bigger part of the image, while larger scales are more detail-oriented. Therefore, the multi-scale supervision, together with the atrous convolution and ASPP, allows our approach to enhance its performance, by having different information of the surrounding area of each of the pixels in the image.

Technically, we are given a set of images, $\{I_1,I_2,\ldots,I_N\}$, with their corresponding semantic segmentation annotations, $\{\phi_1,\phi_2,\ldots,\phi_N\}$. We use the deep network described, and represent $\hat{\theta_i}$ as the score map at the output of our network, and $\hat{\phi_i}$ as our estimation for the semantic segmentation.

The multi-scale learning process solves the following optimization. For each training image, during learning, we apply a soft max function to map the scores given by our network to a probability distribution over the complete set of $C$ classes for each of the $k$ positions in the CNN output map:
\begin{equation}\label{eq:Softmax}
\hat{p}_{kc} = \frac{\exp(\hat{\theta}_{kc})}{\sum_{c'} \exp(\hat{\theta}_{kc'})} \,.
\end{equation}

The result, $\hat{p}_{kc}$, is used to obtain the cross-entropy classification loss, $\zeta$, with which our model is optimized according to the following equation,
\begin{equation}\label{eq:withLoss}
\zeta = \frac{-1}{K} \sum\limits_{k=1}^K \log(\hat{p}_{k},\phi),
\end{equation}
so that our estimation for each pixel in the image, $\hat{\phi}$, will be the class for which the response of the CNN output map is maximum:
\begin{equation}\label{eq:seg_output}
\hat{\phi} = \underset{C}{\operatorname{argmax}} (\hat{\theta}).
\end{equation}

\subsection{Scene Classification}
\label{cls_model}
Since the breakthrough of deep learning and CNNs, scene classification has been carried out using deep networks, originally designed for the object recognition problem, which achieve state-of-the-art results (\eg \cite{He2016,Wang2016,Simonyan2015,Song2017}). On the contrary, in this work, we aim to use the semantic segmentation results to directly classify the scenes. Our results show that our surprisingly simple and fast approach can surpass the complex deep models trained for scene categorization.

Technically, we explore two classifier based solutions, which take as input features our scene labeling output. First, we propose to use the histograms of class labels obtained from the semantic segmentation. Thus, as descriptors for each image, $\bar{F_i}$, we take a $C$-bin histogram, $\operatorname{hist}$, representing the number of pixels of each of the $C$ classes in our semantic segmentation estimation $\hat{\phi}$:
\begin{equation}\label{eq:descriptors}
F_i = \operatorname{hist}(\hat{\phi}) = [f_{i1}, f_{i2},...,f_{iC}],
\end{equation}
that we normalize as follows,
\begin{equation}\label{eq:descriptors_norm}
\bar{F_i} = \operatorname{norm}(F_i),
\end{equation}
where $f_{ic}$ is the number of pixels of class $c$ in image $i$, and $\operatorname{norm}$ corresponds to an L2 normalization. This idea is based on the fact that the objects that appear in an image should define the scene in which they are: the parts \textit{form} the whole. With this histogram-like features, we propose to use SVMs with additive kernels \cite{Vedaldi2012}, because their good performance working with this type of features. In the experiments, we try several of them, such as, intersection kernel, $\chi^2$ or Jensen-Shannon.

On top of that, when constructing the histograms, we also explore the use of a two-level spatial pyramid pooling. With the aim of adding localization information to our descriptors, we divide the image in four parts and extract a histogram, following the procedure described in Equation \ref{eq:descriptors_norm}, for each of them. Thereby, the new descriptors will be a concatenation of five $C$-size vectors: the original one, representing the full image, and one for each of the four second-level parts.

Our second model for scene categorization consists in learning a linear classifier with \textit{one-hot vectors} computed from the semantic segmentation. For each image, we define a \textit{one-hot vector}, $G_i$, as
\begin{equation}\label{eq:One_Shot}
G_i = [u(f_{i1}), u(f_{i2}),...,u(f_{iC})] = [g_{i1}, g_{i2},...,g_{iC}],
\end{equation}
with \textit{u} being the function:
\begin{equation}\label{eq:One_Shot_2}
u(x) = \begin{cases}
1 & x > \delta \\
0 & x \leq \delta
\end{cases}.
\end{equation}

The threshold parameter, $\delta$, is the minimum number of pixels needed to consider that a class appears in the image. Therefore, $g_1,...,g_C \in \{0,1\}$. One-hot vectors just take into account whether a category is present or not in the image. For the categorization, we simply use a linear SVM. The use of one-hot vectors but with the additive kernels do not improve the results, according to our experiments.

\subsection{Object Detection Integration}
\label{obj_model}

Given the semantic segmentation, we present a model to classify the scene, but, can we also solve the object localization task? In this section we describe how to obtain the bounding boxes (BBs) from the pixel labeling masks, in order to evaluate the object detection accuracy of our model, closing, this way, the whole scene understanding problem.

Our solution is embarrassingly simple and fast. Given a semantic segmentation mask, and for each object category $c \in C$, we simply proceed to obtain a binary mask with the pixels that belong to the class of interest. We then trace region boundaries putting a tight BB around each connected component identified in the mask. 

We need to assign a detection score to each of these BBs, and we again rely on the result of the semantic segmentation for this purpose. Given an estimated BB, $BB_i^c$, for the object category $c$, its corresponding object detection score $s_i$ is obtained computing the mean of the confidence scores of the semantic segmentation output pixels within the BB that belong to the object class $c$. In other words, we technically follow this equation:
\begin{equation}\label{eq:obj_detection}
s_i = \frac{1}{P_c^{BB_i^c}} \sum_{j=1}^{BB_i^c}
 \hat{\theta}_{jc},
\end{equation}
where $P_c^{BB_i^c}$ is the total number of pixels of class $c$ in bounding box $BB_i^c$. $ \hat{\theta}_{jc} $ represents the score values of class $c$ of the output map provided by our semantic segmentation network.

\section{EXPERIMENTS}
\label{exp}
To evaluate the effectiveness of our models, we perform scene labeling, scene classification and object detection experiments on the challenging scene understanding NYU Depth V2 dataset \cite{Silberman2012} (NYUD2). The details of the experiments and the corresponding results are provided in the following sections.

\subsection{Experimental Setup}
\label{sec:details}

\subsubsection{Implementation Details}

We start from the publicly available DeepLab model implementation \cite{Chen2016}, which employs the Caffe deep learning framework \cite{Jia2014}. We implement the ResNet-101 \cite{He2016} deep architecture as the base network for the DeepLab model. This network is pretrained on the MS COCO dataset \cite{Tsung2014}. We replace the last layer with a soft max classifier layer, which has as many targets as the number of semantic classes of our task in the NYUD2 dataset. Our loss function is the sum of cross-entropy terms for each spatial position in the CNN output map (subsampled by 8 compared to the original image). All  positions and labels are equally weighted in the overall loss function (except for unlabeled pixels in the ground truth annotations, which are ignored). We optimize the objective function using SGD. For fine tuning, the learning rate is initialized at 0.001 and we apply a \textit{poly} learning rate policy (the learning rate is multiplied by $(1-\frac{iter}{max\_iter})^{power}$) with $power$ parameter fixed to $0.9$. As data augmentation techniques, we apply mirroring, with a probability of 50\%, and cropping: we feed our network with patches of size 385x385 instead of using the complete 480x640 images of the NYUD2 dataset. Our model for semantic segmentation is trained during 20K iterations with a batch size of 1.

With respect to \textit{one-hot vectors} in scene classification, we set the threshold parameter, $\delta$, to 0.5\% of the total number of pixels in the image, since small objects do not tend to be the most representative of a particular type of scene.

\subsubsection{Dataset and Evaluation Metrics}

In our experiments, we use the NYUD2 dataset \cite{Silberman2012}, which consists of 1449 images. We use the publicly available split, which has 795 images for training and 654 images for testing. 

For semantic segmentation, the original task proposed in \cite{Silberman2012} consists of assigning the pixels of the images in just four categories. But, we follow the more complex setup described in \cite{Gupta2013}, and we study a more fine-grained 40 class discrimination task where the scene is divided in \textit{structure categories} like walls, floors, ceiling, windows or doors; \textit{furniture items} like beds, chairs, tables or sofa; and \textit{objects} like lamps, bags, towels or boxes. The complete list is given in Table \ref{table:40_seg}.

We measure the performance of our model in the segmentation task using the pixel accuracy (pixels correctly predicted divided by the total number of pixels annotated), the mean class accuracy (mean of the pixel accuracy when computed for each class separately) and the Jaccard index (true predictions divided by the union of predictions and true labels), which is also known as the intersection over union (IoU).

For the scene categorization task, we follow the experimental setup detailed in \cite{Gupta2013}, where the original 27 categories are reorganized into 10 classes: the 9 most common categories plus the category \textit{other} for the images in the remaining classes. For evaluation, we report the classification accuracy, which is the precision over all test images.

Finally, for the object detection task, we again follow the setup proposed in \cite{Gupta2013}, where 17 of 40 object categories are used. They include items such as toilet, bed, sofa, chair, etc. The ground truth BBs for the objects are obtained from the ground truth semantic segmentations, discarding the ignored pixels (pixels with label 255). We follow the standard PASCAL VOC \cite{voc2012} metric of average precision (AP) for measuring the detection performance. That is, for each object class $c$, we first sort the BBs by their score. Then, we define an overlap criterion of 0.5 between a predicted BB and the ground truth to consider a detection as true positive, and we calculate the precision/recall curve from the ranking obtained. The AP, for each class $c$, is defined as the mean precision at eleven equally spaced recall levels:
\begin{equation}\label{eq:AP}
AP = \frac{1}{11} \sum_{r \in (0,0.1,...,1)} p(r),
\end{equation}
where $p(r)$ is the precision at recall $r$, but interpolated by taking the maximum precision measured for a method for which the corresponding recall exceeds $r$.

\subsection{Scene Labeling}
\begin{figure*}[t]
	\centering
	\vspace*{0.2cm}
	\includegraphics[trim=0cm 15.6cm 0cm 0cm, clip, width=\linewidth]{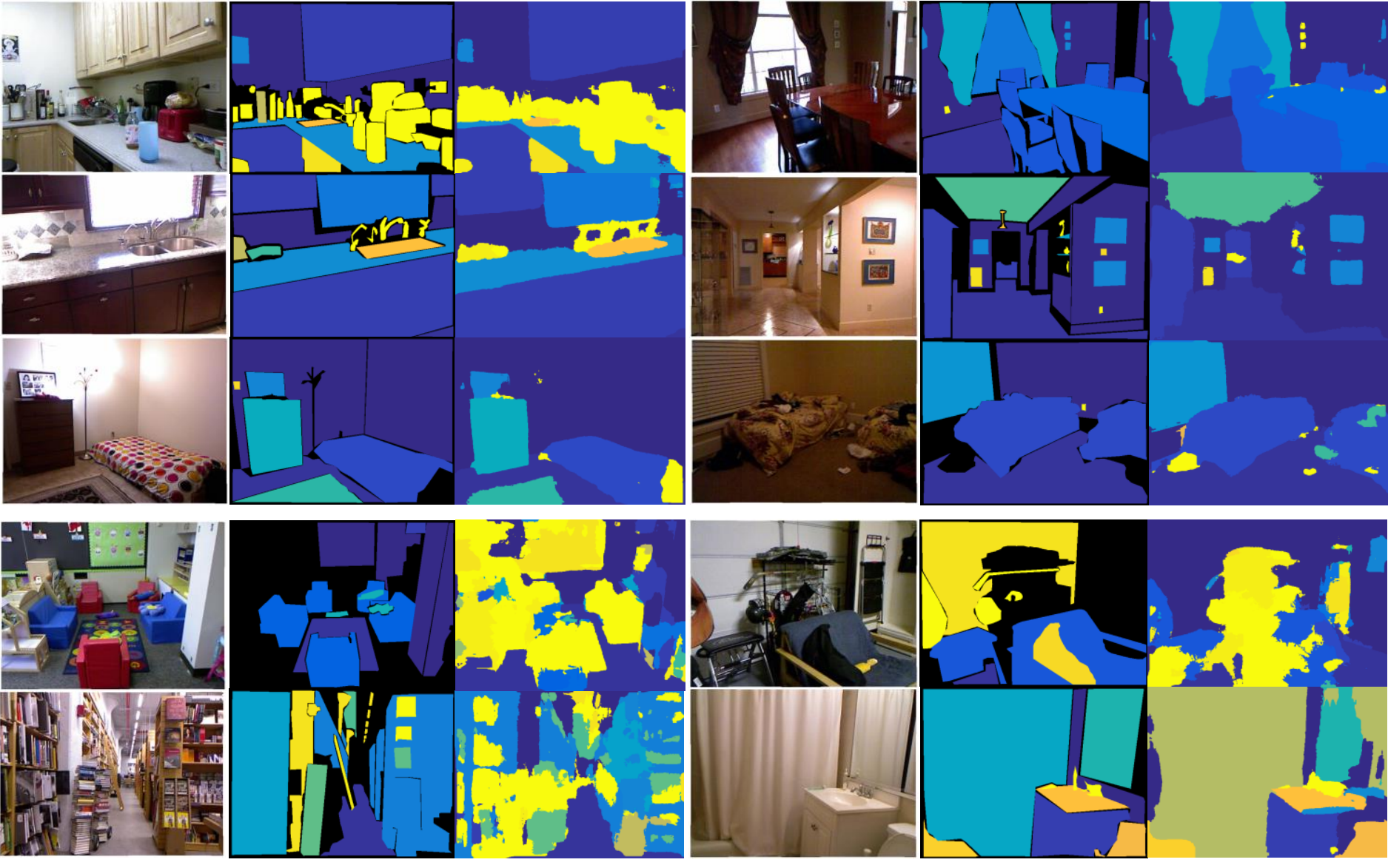}
	\caption {Qualitative results fro the scene labeling problem. For each group of three images, we show, from left to rigth: the RGB image, the ground truth and our results. We represent the best (first 3 rows) and worst (last 2 rows) test images in terms of pixel accuracy.}	
	\label{fig:seg_qualitatives}
\end{figure*}

We here compare our model with other state-of-the-art methods in the task of scene labeling. Several approaches have reported semantic segmentation results on this dataset recently, and some of which make use of depth information to enhance their results. Table \ref{table:seg} shows that our model, simply from RGB images, and using the ResNet-101 as a base network with the aforementioned multi-scale input procedure for the DeepLab architecture, beats them all in all the three evaluation metrics. A set of qualitative results are shown in Figure \ref{fig:seg_qualitatives}.

Finally, in Table \ref{table:40_seg}, we offer detailed IoU results for each of the 40 classes. Our approach outperforms all other methods that have reported these detailed results, obtaining a particularly considerable improvement in the most difficult classes. Therefore, we conclude that the solution proposed for semantic segmentation, defines the new state-of-the-art results for the NYUD2 dataset.

\begin{table}[ht]
	\centering
	\setlength{\tabcolsep}{6pt}
	\renewcommand{\arraystretch}{1.8}
	\caption{Scene labeling performance for the NYUD2 dataset 40-class classification task.}
	\label{table:seg}
	\scalebox{1}{
		\begin{tabular}{  l*{4}{c }}	
			\toprule
			\textbf{Method} & \textbf{Input} & \textbf{pix acc} & \textbf{class acc} & \textbf{IoU} \\ \hline
			RCNN \cite{Gupta2014} & RGB-HHA & 60.3 & 35.1 & 28.6 \\
			FCN-16 \cite{Long2015} & RGB-HHA & 65.4 & 46.1 & 34.0 \\ Eigen et al. \cite{Eigen2014} & RGB-D-N & 65.6 & 45.1 & 34.1 \\
			FuseNet-SF3 \cite{Hazirbas2016} & RGB-D & 66.4 & 44.2 & 34.0 \\ Context-CRF \cite{Lin2016} & RGB & 67.4 & 49.6 & 37.1  \\  MVCNet \cite{Ma2017} & RGB-D & 69.1 & 50.1 & 38.0 \\ 
			\midrule			
			\textbf{Ours} & RGB & \textbf{70.9} & \textbf{52.2} & \textbf{41.8}\\ 
			\bottomrule
	\end{tabular}}
\end{table}

\begin{table*}[ht]
	\setlength{\tabcolsep}{6pt}
	\renewcommand{\arraystretch}{1.6}
	\caption{IoU for each of the classes in the 40-class segmentation task.}
	\label{table:40_seg}
	\begin{center}
		\scalebox{0.583}{
			\begin{tabular}{  l*{20}{c }}	
				\toprule
				& \textbf{wall} & \textbf{floor} & \textbf{cabinet} & \textbf{bed} & \textbf{chair} & \textbf{sofa} & \textbf{table} & \textbf{door} &\textbf{window} & \textbf{book sh.} & \textbf{picture} & \textbf{counter} & \textbf{blinds} & \textbf{desk}& \textbf{shelves}& \textbf{curtain}& \textbf{dresser}& \textbf{pillow}& \textbf{mirror}& \textbf{floor mat}\\ \hline
				
				SC \cite{Silberman2012} & 60.7 & 77.8 & 33.0 & 40.3 & 32.4 & 25.3 & 21.0 & 5.9 & 29.7 & 22.7 & 35.7 & 33.1 & 40.6 & 4.7 & 3.3 & 27.4 & 13.3 & 18.9 & 4.4 & 7.1 \\ \hline
				
				KDES \cite{Ren2012} & 60.0 & 74.4 & 37.1 & 42.2 & 32.5 & 28.2 & 16.6 & 12.9 & 27.7 & 17.3 & 32.4 & 38.6 & 26.5 & 10.1 & 6.1 & 27.6 & 7.0 & 19.7 & 17.9 & 20.1 \\ \hline	

				SVM+Scene \cite{Gupta2013} & 68 & 81 & 48 & 55 & 40 & 44 & 30 & 8.3 & 33 & 20 & 40 & 47 & 44 & 10 & 5.1 & 34 & 22 & 28 & 19 & 22\\ \hline	
				
				RCNN-Depth \cite{Gupta2014} & 68.0 & 81.3 & 44.9 & \textbf{65.0} & 47.9 & 47.9 & 29.9 & 20.3 & 32.6 & 18.1 & 40.3 & 51.3 & 42.0 & 11.3 & 3.5 & 29.1 & 34.8 & 34.4 & 16.4 & 28.0\\ \hline
				
				Det+Scene \cite{Gupta2015} & 67.9 & \textbf{81.5} & 45.0 & 60.1 & 41.3 & 47.6 & 29.5 & 12.9 & 34.8 & 18.1 & 40.7 & 51.7 & 41.2 & 6.7 & 5.2 & 26.9 & 25.0 & 32.8 & 21.2 & \textbf{30.7}\\ \hline
															 	
				\textbf{Ours} & \textbf{74.0} & 80.6 & \textbf{55.9} & 64.6 & \textbf{55.0} & \textbf{61.0} & \textbf{40.9} & \textbf{33.1} & \textbf{45.0} & \textbf{39.9} & \textbf{54.5} & \textbf{53.0} & \textbf{57.1} & \textbf{14.5} & \textbf{10.9} & \textbf{46.6} & \textbf{40.2} & \textbf{37.0} & \textbf{34.0} & 28.4 \\ \hline
				\midrule
				& \textbf{clothes} & \textbf{ceiling} & \textbf{books} & \textbf{fridge} & \textbf{tv} & \textbf{paper} & \textbf{towel} & \textbf{shower cur} & \textbf{box} & \textbf{wh. board} & \textbf{person} & \textbf{night stand} & \textbf{toilet} & \textbf{sink} & \textbf{lamp} & \textbf{bath tub} & \textbf{bag} & \textbf{other str.} & \textbf{other fur.} & \textbf{other prop}  \\ \hline	
				
				SC \cite{Silberman2012} & 6.5 & \textbf{73.2} & 5.5 & 1.4 & 5.7 & 12.7 & 0.1 & 3.6 & 0.1 & 0 & 6.6 & 6.3 & 26.7 & 25.1 & 15.9 & 0 & 0 & 6.4 & 3.8 & 22.4 \\ \hline	
				
				KDES \cite{Ren2012} & 9.5 & 53.9 & 14.8 & 1.9 & 18.6 & 11.7 & 12.6 & 5.4 & 3.3 & 0.2 & 13.6 & 9.2 & 35.2 & 28.9 & 14.2 & 7.8 & 1.2 & 5.7 & 5.5 & 9.7 \\ \hline	

				SVM+Scene \cite{Gupta2013} & 6.9 & 59 & 4.4 & 15 & 9.3 & 1.9 & 14 & 18 & 4.8 & \textbf{37} & 16 & 20 & 50 & 26 & 6.8 & 33 & 0.65 & 6.9 & 2 & 22 \\ \hline
				
				RCNN-Depth \cite{Gupta2014} & 4.7 & 60.5 & 6.4 & 14.5 & 31.0 & 14.3 & 16.3 & 4.2 & 2.1 & 14.2 & 0.2 & 27.2 & 55.1 & 37.5 & 34.8 & \textbf{38.2} & 0.2 & 7.1 & 6.1 & 23.1 \\ \hline
				
				Det+Scene \cite{Gupta2015} & 7.7 & 61.2 & 7.5 & 11.8 & 15.8 & 14.7 & 20.0 & 4.2 & 1.1 & 10.9 & 1.4 & 17.9 & 48.1 & 45.1 & 31.1 & 19.1 & 0.0 & 7.6 & 3.8 & 22.6 \\ \hline

				\textbf{Ours} & \textbf{17.5} & 63.5 & \textbf{29.2} & \textbf{57.5} & \textbf{57.3} & \textbf{27.3} & \textbf{33.9} & \textbf{17.8} & \textbf{9.6} & 32.1 & \textbf{76.9} & \textbf{41.8} & \textbf{73.7} & \textbf{50.2} & \textbf{44.6} & 31.5 & \textbf{7.3} & \textbf{25.6} & \textbf{15.5} & \textbf{32.5} \\				
				\bottomrule  
		\end{tabular}}
	\end{center}
\end{table*}

\subsection{Scene Classification}

Table \ref{table:sota_cls} shows the results of our methods when we compare them with other state-of-the-art models for scene recognition in the NYUD2 dataset. We incorporate to the evaluation, two baselines methods, which are a VGG-16 \cite{Simonyan2015} and a ResNet-101 \cite{He2016} architecture, which we train to directly classify the scenes. Technically, we develop a fine tuning procedure for two pretrained models of these networks, where the objective now is correctly classify the scenes in the NYUD2 dataset. 

Using the proposed 40-classes scene labeling histograms as features to feed a linear SVM, already improves all previously reported methods, and stays very close to the ResNet-101 baseline model, specifically trained for the problem of scene classification. If instead we use one-hot vectors as described in Section \ref{cls_model}, we increase the accuracy by 1,3\%. Using a Jensen-Shannon kernel for our SVM, with normalized histograms, obtains another 0,3\%. Finally, applying a two-level spatial pyramid pooling technique improves the performance by an extra 0,76\%, attaining a 68,96\% classification accuracy, the new state-of-the-art for this dataset.

\begin{table}[t]
	\centering
	\vspace*{0.25cm}
	\setlength{\tabcolsep}{6pt}
	\renewcommand{\arraystretch}{1.5}
	\caption{Scene classification accuracy for NYUD2. Comparison with the state of the art.}
	\label{table:sota_cls}
	\scalebox{0.7}{
		\begin{tabular}{  c*{3}{c }}	
			\toprule
			& \textbf{Method} & \textbf{Input} & \textbf{Acc.} \\ \hline
			\multirow{3}{*}{State of the art} & RCNN-Depth \cite{Gupta2014} & RGB-D & 45.4 \\
			& FV+CNN \cite{Wang2016} & RGB-D & 63.9 \\ 
			& D-CNN \cite{Song2017} & RGB-D & 65.8 \\
			\midrule
			\multirow{2}{*}{Baselines} & VGG-16 classification & RGB & 64.37 \\
			& ResNet-101 classification & RGB & 67.73 \\ 
			\midrule
			\multirow{4}{*}{\textbf{Ours}} & \textbf{Histograms + Linear SVM} & RGB & 66.51\\
			& \textbf{One-hot + Linear SVM} & RGB & 67.89\\ 
			& \textbf{Histograms + Additive Kernel SVM} & RGB & 68.20\\ 
			& \textbf{Histograms + Spatial Pyramid + Additive Kernel SVM} & RGB & \textbf{68.96}\\ 
			\bottomrule
	\end{tabular}}
\end{table}

\subsection{Object Detection}
\label{obj_det_res}

Here, we are interested in investigating the task of detecting furniture like objects in indoor scenes. The NYUD2 dataset \cite{Silberman2012} was originally proposed to study semantic segmentation and scene recognition. However, we can derive BBs annotations for the objects directly from the semantic segmentation ground truth, following the experimental setup of \cite{Gupta2013}. Hence, we can also use this dataset to evaluate the performance of the models in an object localization task.

\begin{figure*}[t]
\centering
\vspace*{0.2cm}
\includegraphics[width=0.9\linewidth]{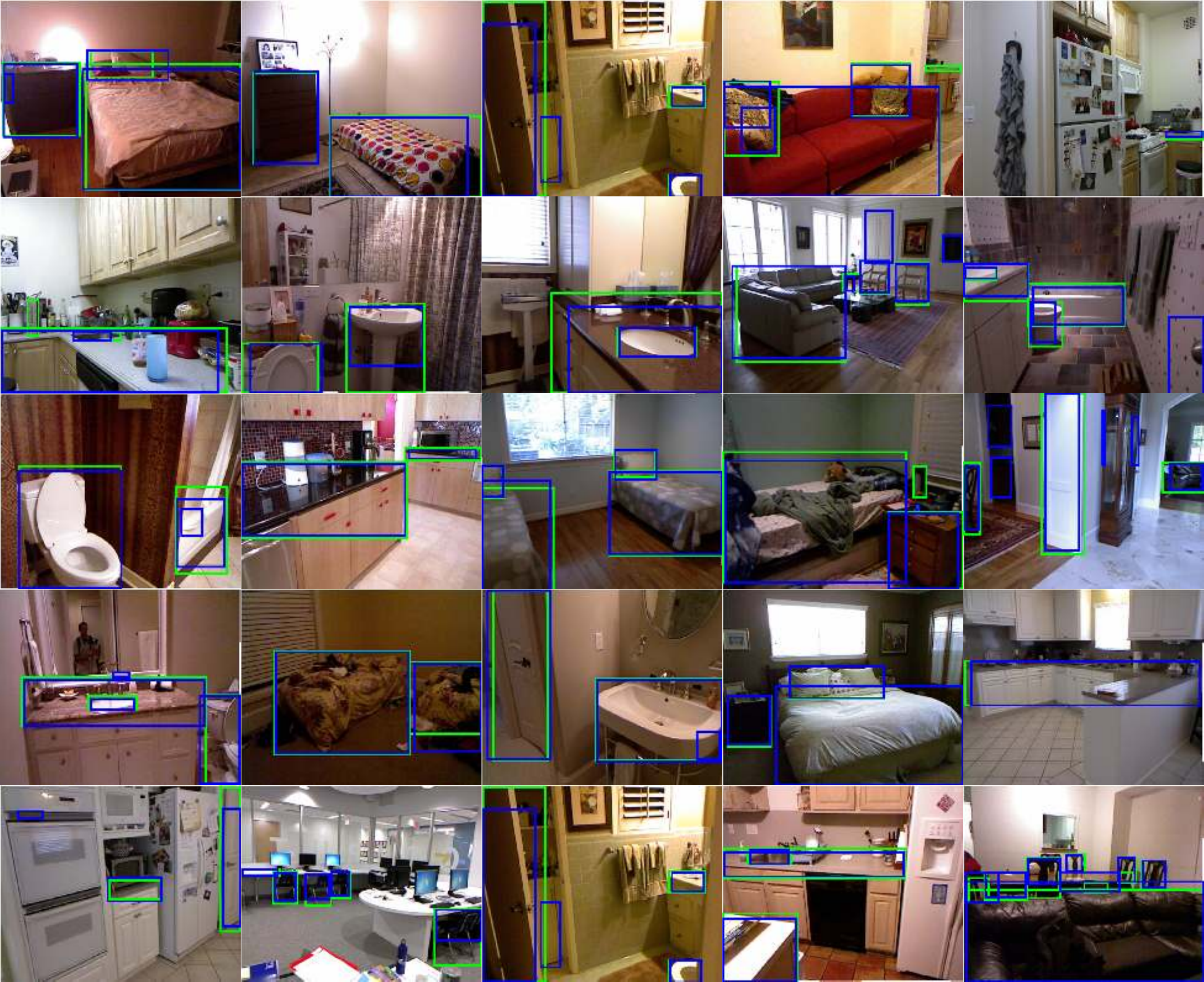}
\caption{Qualitative Results for the Object Detection Task. The ground truth bounding boxes are shown in green and our predicted bounding boxes in blue.}
\label{fig:qualitative_det}
\end{figure*}

Table \ref{table:sota_det} summarizes the results achieved by our simple object localization solution and the proposed approaches described by Gupta \etal in \cite{Gupta2013,Gupta2015,Gupta2014}. We observe that we are able to consistently outperform all these previous works. Note that our simple approach, using only RGB images, outperforms all these methods which jointly exploit the depth and appearance information. Importantly, we achieve an improvement of more than $5\%$ for the AP compared to the best results detailed in \cite{Gupta2014}. We provide some qualitative detection results in Figure \ref{fig:qualitative_det}.

\begin{table}[t]
\caption{Object Detection Performance, using the AP metric, for the NYUD2 dataset. Comparison with the state of the art.}
\label{table:sota_det}
\begin{center}
\scalebox{0.75}{
\begin{tabular}{l|cccc}
\toprule
\multicolumn{1}{c|}{\textbf{}} & \multicolumn{1}{c}{SVM+Scene \cite{Gupta2013}} & \multicolumn{1}{c}{Det+Scene \cite{Gupta2015}} & \multicolumn{1}{c}{RCNN-Depth \cite{Gupta2014}} & \multicolumn{1}{c}{\textbf{Ours}}\\
\multicolumn{1}{c|}{Input} & \multicolumn{1}{c}{\textbf{RGBD}} & \multicolumn{1}{c}{\textbf{RGBD}} & \multicolumn{1}{c}{\textbf{RGBD}} & \multicolumn{1}{c}{\textbf{RGB}}\\
\midrule
Bed & 52.1 & 56.0 & \textbf{66.5} & 58.7\\
Chair & 6.4 & 23.5 & 40.8 & \textbf{43.9}\\
Sofa  & 17.5 & 34.2 & 42.8 & \textbf{52.7}\\
Counter & 32.7 & 24.0 & 37.6 & \textbf{51.3}\\
Lamp & 1.4 & 26.7 & \textbf{29.3} & 27.7\\
Pillow & 3.3 & 20.7 & \textbf{37.4} & 19.2\\
Sink & 14.0 & 22.8 & 24.2 & \textbf{46.1}\\
Table & 9.3 & 17.2 & \textbf{24.3} & 23.4\\
Bathtub & 28.4 & 19.3 & 22.9 & \textbf{33.7}\\
Television & 3.1 & 19.5 & 37.2 & \textbf{42.7}\\
Bookshelf & 6.7 & 17.5 & 21.8 & \textbf{35.1}\\
Toilet & 13.3 & 41.5 & 53.0 & \textbf{74.0}\\
Box & 0.7 & 0.6 & \textbf{3.0} & 2.5\\
Desk & 0.8 & 6.2 & \textbf{10.2} & 7.0\\
Door & 5.0 & 9.5 & 20.5 & \textbf{23.3}\\
Dresser & 13.3 & 16.4 & 26.2 & \textbf{26.5}\\
Night-stand & -- & 32.6 & 39.5 & \textbf{43.5}\\ 
\midrule
Mean over 16 classes & 13 & 24.3 & 31.1 & \textbf{35.5}\\
Mean over 17 classes & -- & 23.0 & 31.6 & \textbf{36.0}\\
\bottomrule
\end{tabular}
}
\end{center}
\end{table}

\section{CONCLUSION}
\label{conclusion}
This paper proposes a highly effective \textit{bottom-up} approach to perform scene understanding. From an accurate multi-scale semantic segmentation deep network, using the DeepLab model with ResNets, we present and develop a variety of simple but robust strategies to use scene labeling-based descriptors for performing both scene classification and object detection tasks. 

We compare our solutions with other state-of-the-art methods using a variety of metrics in the challenging NYUD2 dataset. The obtained results confirm that our methods are able to outperform \emph{all} other previously reported models in all the three tasks comprising scene understanding, by only using RGB images. 

Overall, in this work we demonstrate that to rely on a precise scene labeling, as the one given by our model, for acquiring a complete scene understanding, is an extremely effective solution. Therefore, \emph{in pixels we trust}.

{\small
\bibliographystyle{IEEEtran}
\bibliography{IEEEexample}
}

\end{document}